\documentclass{article}

\usepackage[utf8]{inputenc}
\usepackage{arxiv}

\usepackage{amsmath,amssymb,amsfonts}
\usepackage{graphicx}
\usepackage{booktabs}
\usepackage{siunitx}
\usepackage{hyperref}
\usepackage{caption}
\usepackage{subcaption}
\usepackage{listings}
\usepackage{xcolor}

\lstdefinestyle{mypython}{
    language=Python,
    basicstyle=\ttfamily\small,
    keywordstyle=\color{blue}\bfseries,
    commentstyle=\color{gray}\itshape,
    stringstyle=\color{orange},
    showstringspaces=false,
    breaklines=true,
    frame=single,
    columns=fullflexible
}
\title{Sarcasm Detection on Reddit Using Classical Machine Learning and Feature Engineering}

\author{
  Subrata Karmaker \\
  Department of Mathematics\\
  Technische Universit\"at Chemnitz\\
  Germany\\
  \texttt{skarmaker.tuc@gmail.com}
}

\begin{document}
\maketitle

\begin{abstract}
Sarcasm is something people recognise instinctively, often without even noticing how they do it. Online spaces like Reddit are full of short comments that can sound sincere or sharply ironic depending on the context, tone and sometimes even on who is speaking. Machines, of course, do not have access to any of this subtlety. A small sentence such as ``Great job.'' might be encouragement or criticism, and without clues from the surrounding conversation, it is genuinely ambiguous.

In this study, I explore whether a set of straightforward, older machine learning methods---methods that many people now overlook in favour of neural networks---can still pick up enough textual cues to separate sarcastic comments from straightforward ones. I work with a 100\,000-comment subsample of the Self-Annotated Reddit Corpus (SARC\,2.0) and restrict myself intentionally to replies alone, without parent comments. To represent each reply, I combine word-based and character-based TF--IDF features with a few simple stylistic measurements such as length, uppercase ratios and punctuation usage. On top of this feature space, I train and evaluate logistic regression, a linear SVM, multinomial Naive Bayes and a random forest.

Although the performance is limited by the lack of conversational context, the results are surprisingly steady. Logistic regression and Naive Bayes reach $F_{1}$-scores of around $0.57$ for identifying sarcastic comments. The performance does not rival contextual or neural models, but the goal here is not to compete with them. Instead, it is to offer a clean, reproducible baseline for what can be achieved with transparent, lightweight methods. All code used for this project is included at the end of the paper.
\end{abstract}

\section{Background and Motivation}

Anyone who spends time on Reddit knows how common sarcasm is there. People often use it to soften disagreement, to joke about something unpleasant or simply out of habit. What is obvious to human readers becomes surprisingly difficult for machines. Sarcasm rarely announces itself; rather, its meaning relies on subtle contrasts, expectations and prior statements. Even as a human, reading a reply without knowing the context can sometimes leave one unsure whether the writer was being serious or ironic.

When building automatic classifiers, we often hope that an algorithm can infer sentiment or intent from the words alone. With sarcasm, that hope quickly runs into obstacles. The intended meaning often contradicts the literal phrasing. For example, ``Wonderful idea'' might be completely earnest or bitterly sarcastic. Without the earlier part of the conversation, the two interpretations are indistinguishable.

Despite these complications, it remains interesting---and sometimes practically important---to understand how far we can get with models that have access only to the text at hand. Classical machine learning models, the kind that were common before neural networks became dominant, still have advantages in interpretability and computational efficiency. They train quickly, and their behaviour is easier to analyse, which makes them appealing as baselines. The intention behind this study is not to claim they solve sarcasm but to see what signals they capture, even when the task is intentionally stripped of context.

This project grew partly out of curiosity and partly out of a desire to document a baseline that others can build on. Especially now, as deep learning architectures become more complex, it is useful to understand what simpler models can do and where they fall short. This clarity helps highlight what context or deeper representation actually adds.

\section{Connections to Prior Research}

Researchers have been thinking about sarcasm detection for quite a while. Joshi and colleagues surveyed many of the early attempts and pointed out how difficult the problem is when one relies mainly on surface-level lexical cues \cite{Joshi2017Survey}. Much of that work used techniques similar in spirit to what I attempt here: extracting handcrafted features and feeding them into standard classifiers. These methods showed that certain patterns, such as exaggerated sentiment or particular word combinations, tend to correlate with sarcasm, but the overall performance was naturally limited.

A notable shift occurred when researchers began to incorporate surrounding conversational context. Ghosh et al.\ examined how replies relate to the messages they respond to and found that context often turns an ambiguous comment into an unmistakably sarcastic one \cite{Ghosh2017Context}. This idea makes intuitive sense; sarcasm frequently depends on knowing what came beforehand. Without it, one is left guessing.

More recently, studies have explored multimodal settings. Some approaches integrate images, audio or user profiles into the prediction pipeline. Farabi and collaborators discuss how combining multiple sources of information can uncover sarcastic intent that plain text alone fails to reveal \cite{Farabi2024Multimodal}. These systems tend to be more complex, but they reflect the reality that sarcasm in the wild is rarely expressed in text alone.

A recurring dataset in many of these studies is the Self-Annotated Reddit Corpus (SARC), introduced by Khodak et al.\ \cite{Khodak2018SARC}. It is a valuable resource precisely because it is large and user-annotated, and because it contains both the replies and their parent comments. In the present study, I intentionally ignore the parent comments. The motivation for this is not because context is unimportant, but because leaving it out helps create a controlled and transparent baseline that future researchers can compare against when introducing more advanced methods.

\section{Data and Preparation}

The data used in this work come from the balanced subset of SARC\,2.0. Each entry includes a label indicating whether the reply is sarcastic, metadata about the users and conversation, the parent comment and finally the reply text itself. For this study, I only keep the binary label and the reply text. This constraint forces the model to rely entirely on the wording and style of the reply, with no contextual clues.

Since the full balanced subset is quite large, I draw a subsample of 100\,000 entries. While this reduces coverage, it makes experimentation more manageable, especially when trying different feature designs or classifier settings. The subsampling uses a fixed random seed to ensure that results are reproducible. I also remove replies that are empty or consist only of whitespace, because they add noise without contributing information.

After cleaning, I use an 80/20 train--test split, stratified so that the proportion of sarcastic and non-sarcastic comments remains stable across both sets. Using stratification is important here because sarcasm is evenly balanced in the dataset, and a careless split might distort that balance. The test set is held aside and only used after all models have been trained.

\section{Feature Design and Modelling Strategy}

Representing the text in a way that captures both content and style is one of the most critical parts of this project. I created a custom transformer, which I call \texttt{TextFeatures}, to assemble different kinds of information into a single representation.

The first component is a word-based TF--IDF representation using unigrams and bigrams. This captures which words and short phrases appear in a reply and how characteristic they are compared with the rest of the dataset. Many sarcastic comments carry specific lexical markers, and TF--IDF helps highlight these.

Alongside the word-level features, I include a character-based TF--IDF representation. By focusing on character sequences of length three to five, this representation becomes sensitive to stylistic nuances such as elongated words, creative spellings or distinctive punctuation patterns. These subtle features sometimes carry more information about sarcasm than the literal words themselves.

Beyond TF--IDF, I add a handful of numeric indicators that aim to summarise the style of writing. These include the length of the reply, the number of words, the ratio of uppercase letters and the frequency of exclamation and question marks adjusted by word count. They are simple features, but while working with the data, I noticed that many sarcastic replies rely on a certain kind of exaggeration that these measures can help quantify.

All of these components---word features, character features and the numeric indicators---are joined into a single sparse feature matrix. Each classifier then receives exactly the same representation. I experiment with four classical models: logistic regression, a linear support vector machine, multinomial Naive Bayes and a random forest. Logistic regression and SVMs are common choices in text classification because they work well in high-dimensional spaces. Naive Bayes, although based on simplifying assumptions, often performs surprisingly well on TF--IDF features. The random forest introduces a non-linear alternative that might capture interactions between features that linear models cannot.

Each model is wrapped in a scikit-learn \texttt{Pipeline}, which ensures that the feature extraction is always applied in the same way. This uniformity avoids subtle differences in preprocessing that could distort the comparison.

\section{Experimental Findings}

Evaluating sarcasm detection requires more than simply measuring accuracy, because accuracy alone does not show how well a model identifies sarcastic comments specifically. In addition to accuracy, I therefore look closely at precision, recall and the $F_{1}$-score for the sarcastic class.

The four models perform at broadly similar levels, with Naive Bayes and logistic regression showing almost identical performance. The $F_{1}$-scores for sarcasm hover around $0.57$, which is modest but meaningful in a context-free setting. The linear SVM performs slightly worse, and the random forest does not show a clear advantage over the linear models. Table~\ref{tab:main-results} summarises the key results.

\begin{table}[h]
\centering
\caption{Performance of classical models on the SARC balanced subset (100k subsample). Metrics refer to the sarcastic class.}
\label{tab:main-results}
\begin{tabular}{lcccc}
\toprule
Model & Accuracy & Precision & Recall & $F_1$ \\
\midrule
Logistic Regression & 0.564 & 0.564 & 0.574 & 0.569 \\
Linear SVM          & 0.541 & 0.542 & 0.534 & 0.538 \\
Naive Bayes         & \textbf{0.565} & 0.566 & \textbf{0.574} & \textbf{0.569} \\
Random Forest       & 0.558 & 0.558 & 0.568 & 0.563 \\
\bottomrule
\end{tabular}
\end{table}

\begin{figure}[h]
\centering
\includegraphics[width=0.55\textwidth]{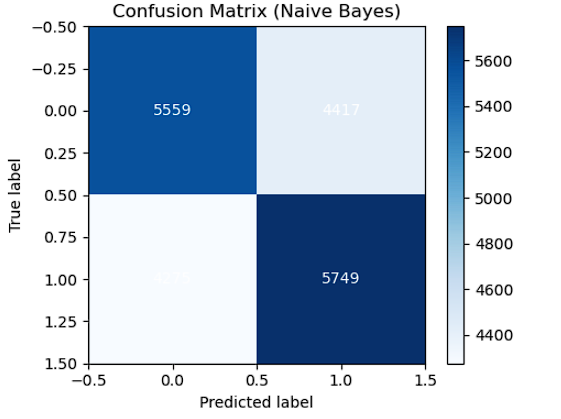}
\caption{Confusion matrix for the multinomial Naive Bayes classifier.}
\label{fig:cm-nb}
\end{figure}

To understand where the models succeed and where they fail, I examined the confusion matrix for the Naive Bayes classifier. The matrix, shown in Figure~\ref{fig:cm-nb}, reveals that a considerable number of sarcastic comments are recognised correctly, but many are still misclassified as non-sarcastic. The reverse is also true: some non-sarcastic replies are interpreted as sarcastic. This back-and-forth pattern reflects the inherent ambiguity of the task.

\begin{figure}[h]
\centering
\includegraphics[width=0.55\textwidth]{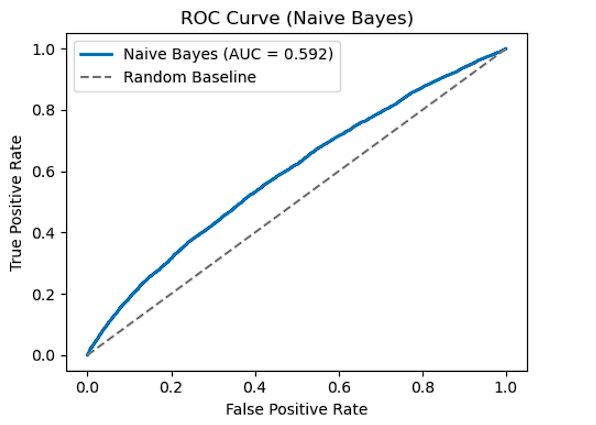}
\caption{ROC curve for the multinomial Naive Bayes classifier (solid line) compared with a random baseline (dashed line).}
\label{fig:roc-nb}
\end{figure}

Looking at the ROC curve gives a more holistic view of the classifier’s discrimination ability. As depicted in Figure~\ref{fig:roc-nb}, the curve sits noticeably above the diagonal line of a random baseline. The area under the curve is about $0.59$. This indicates that while the model does better than chance, there is still substantial overlap between the two classes in the feature space.

\section{Reflections and Interpretation}

As I analysed the results, it became increasingly clear that sarcasm detection without context is fundamentally difficult. The models do find patterns---characteristics of writing style and particular word combinations---that help them distinguish sarcastic replies from literal ones. Yet the limitations are just as obvious. Many replies are simply too ambiguous in isolation, and even a human reader might hesitate without seeing the parent comment.

One interesting aspect I noticed is that the models’ strengths seem to come from different kinds of signals. The TF--IDF features capture lexical tendencies, while the numeric measures highlight stylistic exaggeration. Naive Bayes appears to balance these reasonably well, which might explain why it performs slightly better than the others. Its probabilistic structure seems particularly well suited for sparse TF--IDF data.

The modest performance also reflects the nature of online language. People on Reddit often write loosely, using in-jokes, abbreviations or cultural references that the model has no way to interpret. Even when the surface text looks sarcastic, the intended tone might depend heavily on the earlier message or on the relationship between the participants. Without that information, the classifier can only guess based on what is literally written.

Despite these limitations, I find value in the clarity offered by classical models and carefully engineered features. The behaviour of these models is easier to understand and diagnose than that of modern neural architectures. For researchers who wish to build more sophisticated sarcasm detectors, this study can serve as a reference point: a baseline that makes explicit what is achievable without context or deep representations.

\section{Concluding Remarks}

This work set out to examine how much can be achieved in sarcasm detection using classical machine learning techniques and a deliberately constrained feature space. By focusing exclusively on reply text and combining TF--IDF representations with simple stylistic features, I trained four widely used classifiers on a balanced subset of SARC\,2.0. The results show that logistic regression and Naive Bayes reach $F_{1}$-scores around $0.57$ for sarcastic comments, recognising some meaningful patterns but inevitably missing many subtleties.

The purpose of the study was not to achieve the best possible performance but to create a transparent and reproducible baseline. The full Python code is provided in the appendix, allowing others to retrace each step or extend the analysis. In future research, one could incorporate conversation context, explore pre-trained neural embeddings or experiment with larger and more diverse datasets. These directions may shed further light on how much of sarcasm is expressed through style alone and how much depends on deeper layers of meaning.

\section*{Acknowledgements}

I would like to thank the creators of the SARC dataset for making it publicly available. Their contribution makes studies like this possible. I am also grateful for the broader research community whose work on text classification and conversational analysis continues to inspire new approaches and ideas.

\bibliographystyle{plain}
\bibliography{refs}

@article{Joshi2017Survey,
  author    = {Aditya Joshi and Pushpak Bhattacharyya and Mark J. Carman},
  title     = {Automatic Sarcasm Detection: A Survey},
  journal   = {ACM Computing Surveys},
  year      = {2017},
  volume    = {50},
  number    = {5},
  pages     = {1--22},
  doi       = {10.1145/3132687}
}

@article{Ghosh2017Context,
  author    = {Debanjan Ghosh and Alexander R. Fabbri and Smaranda Muresan},
  title     = {Sarcasm Analysis Using Conversation Context},
  journal   = {Computational Linguistics},
  year      = {2017},
  volume    = {43},
  number    = {3},
  pages     = {755--792},
  doi       = {10.1162/COLI_a_00302}
}

@inproceedings{Farabi2024Multimodal,
  author    = {Shahid Farabi and Ana L{\'o}pez and Nadir Durrani and Fahad AlObaidly},
  title     = {A Survey of Multimodal Sarcasm Detection},
  booktitle = {Proceedings of the International Joint Conference on Artificial Intelligence (IJCAI)},
  year      = {2024}
}

@inproceedings{Khodak2018SARC,
  author    = {Mikhail Khodak and Nikunj Saunshi and Kiran Vodrahalli},
  title     = {A Large Self-Annotated Corpus for Sarcasm},
  booktitle = {Proceedings of the Eleventh International Conference on Language Resources and Evaluation (LREC)},
  year      = {2018}
}

\appendix
\section{Full Python Code}
\label{sec:code}

\vspace{1em}
\noindent
The code below contains the full implementation used in all experiments. It includes data loading, feature extraction, model training and the generation of evaluation plots.

\begin{lstlisting}[style=mypython]
from typing import List, Optional
from pathlib import Path

import numpy as np
import pandas as pd

from scipy.sparse import hstack, csr_matrix

from sklearn.base import BaseEstimator, TransformerMixin
from sklearn.feature_extraction.text import TfidfVectorizer

from sklearn.linear_model import LogisticRegression
from sklearn.naive_bayes import MultinomialNB
from sklearn.svm import LinearSVC
from sklearn.ensemble import RandomForestClassifier
from sklearn.pipeline import Pipeline
from sklearn.model_selection import train_test_split
from sklearn.metrics import (
    accuracy_score,
    classification_report,
    confusion_matrix,
    roc_auc_score,
    roc_curve,
)

import matplotlib.pyplot as plt


class TextFeatures(BaseEstimator, TransformerMixin):
    """
    Custom feature engineering transformer for sarcasm detection.
    It combines:
    - word-level TF-IDF features
    - character-level TF-IDF features
    - simple numeric stylistic features
    """

    def __init__(
        self,
        max_features_word: int = 20000,
        max_features_char: int = 10000,
        ngram_range_word=(1, 2),
        ngram_range_char=(3, 5),
        lowercase: bool = True,
        stop_words: Optional[str] = "english",
    ):
        self.max_features_word = max_features_word
        self.max_features_char = max_features_char
        self.ngram_range_word = ngram_range_word
        self.ngram_range_char = ngram_range_char
        self.lowercase = lowercase
        self.stop_words = stop_words

        self.word_vectorizer_: Optional[TfidfVectorizer] = None
        self.char_vectorizer_: Optional[TfidfVectorizer] = None

    def _basic_numeric_features(self, texts: List[str]) -> np.ndarray:
        """
        Compute simple numeric features:
        - length in characters
        - number of words
        - exclamation marks per word
        - question marks per word
        - uppercase ratio
        """
        lengths = np.array([len(t) for t in texts], dtype=float)
        num_words = np.array([len(t.split()) for t in texts], dtype=float) + 1.0
        num_exclam = np.array([t.count("!") for t in texts], dtype=float)
        num_question = np.array([t.count("?") for t in texts], dtype=float)
        num_upper = np.array(
            [sum(1 for ch in t if ch.isupper()) for t in texts],
            dtype=float,
        )

        exclam_per_word = num_exclam / num_words
        question_per_word = num_question / num_words
        upper_ratio = num_upper / lengths.clip(min=1.0)

        features = np.vstack(
            [lengths, num_words, exclam_per_word, question_per_word, upper_ratio]
        ).T
        return features

    def fit(self, X, y=None):
        """
        Fit word and character TF-IDF vectorizers on the training texts.
        """
        if isinstance(X, pd.Series):
            texts = X.astype(str).tolist()
        elif isinstance(X, (list, np.ndarray)):
            texts = [str(t) for t in X]
        else:
            raise ValueError(f"Unsupported input type for TextFeatures: {type(X)}")

        self.word_vectorizer_ = TfidfVectorizer(
            max_features=self.max_features_word,
            ngram_range=self.ngram_range_word,
            lowercase=self.lowercase,
            stop_words=self.stop_words,
            sublinear_tf=True,
        )

        self.char_vectorizer_ = TfidfVectorizer(
            max_features=self.max_features_char,
            ngram_range=self.ngram_range_char,
            lowercase=self.lowercase,
            analyzer="char",
            sublinear_tf=True,
        )

        self.word_vectorizer_.fit(texts)
        self.char_vectorizer_.fit(texts)
        return self

    def transform(self, X):
        """
        Transform texts into a combined sparse feature matrix.
        """
        if isinstance(X, pd.Series):
            texts = X.astype(str).tolist()
        elif isinstance(X, (list, np.ndarray)):
            texts = [str(t) for t in X]
        else:
            raise ValueError(f"Unsupported input type for TextFeatures: {type(X)}")

        word_tfidf = self.word_vectorizer_.transform(texts)
        char_tfidf = self.char_vectorizer_.transform(texts)
        dense_feats = self._basic_numeric_features(texts)

        dense_sparse = csr_matrix(dense_feats.astype(float))
        return hstack([word_tfidf, char_tfidf, dense_sparse])


def make_logreg_model() -> Pipeline:
    """
    Logistic regression pipeline.
    """
    return Pipeline(
        [
            ("features", TextFeatures()),
            ("clf", LogisticRegression(max_iter=500, n_jobs=-1)),
        ]
    )


def make_svm_model() -> Pipeline:
    """
    Linear SVM pipeline.
    """
    return Pipeline(
        [
            ("features", TextFeatures()),
            ("clf", LinearSVC(max_iter=5000, dual="auto")),
        ]
    )


def make_nb_model() -> Pipeline:
    """
    Multinomial Naive Bayes pipeline.
    """
    return Pipeline(
        [
            ("features", TextFeatures()),
            ("clf", MultinomialNB()),
        ]
    )


def make_rf_model() -> Pipeline:
    """
    Random Forest pipeline.
    """
    return Pipeline(
        [
            ("features", TextFeatures()),
            ("clf", RandomForestClassifier(
                n_estimators=150,
                max_depth=None,
                n_jobs=-1,
                random_state=42,
            )),
        ]
    )


def get_all_models():
    """
    Return a dictionary of all models to be evaluated.
    """
    return {
        "logreg": make_logreg_model(),
        "svm": make_svm_model(),
        "nb": make_nb_model(),
        "rf": make_rf_model(),
    }


# =======================
# Main experiment script
# =======================

RANDOM_STATE = 42
data_path = Path("train-balanced.csv.bz2")

# Load the balanced SARC subset
df = pd.read_csv(
    data_path,
    compression="bz2",
    sep="\t",
    header=None,
    engine="python",
    quoting=3,
    on_bad_lines="skip",
)

# Subsample for computational efficiency
df_small = df.sample(n=100000, random_state=RANDOM_STATE)

# Label and reply text
y = df_small.iloc[:, 0].astype(int)
X = df_small.iloc[:, 9].astype(str)

# Remove empty / whitespace-only texts
mask = X.str.strip().astype(bool)
X = X[mask]
y = y[mask]

# Stratified train-test split
X_train, X_test, y_train, y_test = train_test_split(
    X,
    y,
    test_size=0.2,
    random_state=RANDOM_STATE,
    stratify=y,
)

models = get_all_models()
metrics_summary = {}

# Train and evaluate each model
for name, model in models.items():
    print("=" * 60)
    print(f"Training model: {name}")
    model.fit(X_train, y_train)
    y_pred = model.predict(X_test)

    acc = accuracy_score(y_test, y_pred)
    report_dict = classification_report(
        y_test, y_pred, digits=3, output_dict=True
    )

    # Print detailed classification report
    print("Accuracy:", acc)
    print(classification_report(y_test, y_pred, digits=3))

    # Store key metrics for the sarcastic class (label "1")
    metrics_summary[name] = {
        "accuracy": acc,
        "precision_sarcastic": report_dict["1"]["precision"],
        "recall_sarcastic": report_dict["1"]["recall"],
        "f1_sarcastic": report_dict["1"]["f1-score"],
    }

# Summary table for all models
metrics_df = pd.DataFrame(metrics_summary).T
print("\nSummary metrics:")
print(metrics_df)

# ========== Confusion Matrix & ROC Curve for Naive Bayes ==========

nb_model = models["nb"]
y_pred_nb = nb_model.predict(X_test)

# Confusion matrix
cm = confusion_matrix(y_test, y_pred_nb)
print("\nConfusion matrix (Naive Bayes):")
print(cm)

plt.figure(figsize=(5, 4))
plt.imshow(cm, cmap="Blues")
plt.title("Confusion Matrix (Naive Bayes)")
plt.xlabel("Predicted label")
plt.ylabel("True label")
plt.colorbar()

# Add counts inside the cells
for i in range(cm.shape[0]):
    for j in range(cm.shape[1]):
        plt.text(
            j, i, cm[i, j],
            ha="center",
            va="center",
            color="white" if cm[i, j] > cm.max() / 2.0 else "black",
        )

plt.tight_layout()
plt.show()
plt.savefig("confusion_matrix_nb.png", dpi=300, bbox_inches="tight")
plt.close()

# ROC curve for Naive Bayes
y_score_nb = nb_model.predict_proba(X_test)[:, 1]
fpr, tpr, _ = roc_curve(y_test, y_score_nb)
auc = roc_auc_score(y_test, y_score_nb)
print(f"\nAUC (Naive Bayes): {auc:.3f}")

plt.figure(figsize=(5, 4))

# Model ROC line
plt.plot(fpr, tpr, label=f"Naive Bayes (AUC = {auc:.3f})", linewidth=2)

# Random baseline line
plt.plot([0, 1], [0, 1], "--", color="gray", label="Random baseline")

plt.xlabel("False Positive Rate")
plt.ylabel("True Positive Rate")
plt.title("ROC Curve (Naive Bayes)")
plt.legend()

plt.tight_layout()
plt.show()
plt.savefig("roc_curve_nb.png", dpi=300, bbox_inches="tight")
plt.close()
\end{lstlisting}

\end{document}